%% file: naacl2021.tex
\pdfoutput=1
\documentclass[11pt]{article}
\usepackage{naacl2021}
\usepackage[T1]{fontenc}
\usepackage{microtype}

\usepackage{times}
\usepackage{latexsym}
\usepackage{enumitem}
\usepackage{paralist}
\usepackage{caption}
\usepackage{subcaption}

\usepackage{comment}
\usepackage[flushleft]{threeparttable}
\usepackage{graphicx}
\usepackage{amsmath}
\usepackage{tabularx}
\usepackage{mathrsfs}
\usepackage{multirow}
\usepackage{algorithm}
\usepackage{enumitem}
\usepackage{xcolor}
\usepackage{listings}
\usepackage{hyperref}

\usepackage[frozencache,cachedir=.]{minted}
\usepackage{amssymb}
\usepackage{pifont}

\usepackage[utf8]{inputenc}
\usepackage{cleveref}
\usepackage{minted}
\crefname{section}{§}{§§}
\Crefname{section}{§}{§§}

\usepackage{xcolor}

\definecolor{MyColor}{RGB}{50, 100, 250}
\definecolor{Orange}{RGB}{244, 101, 66}
\definecolor{Red}{RGB}{255, 0, 0}
\definecolor{Green}{RGB}{0, 255, 0}
\definecolor{Blue}{RGB}{0, 0, 255}

\newcommand{\tool}{PLBART~}
\newcommand{\eg}{\emph{e.g.,}~}
\newcommand{\ie}{\emph{i.e.,}~}
\newcommand{\wrt}{\emph{w.r.t.}~}

\usepackage{arydshln}
\usepackage[belowskip=0pt,aboveskip=5pt]{caption}
\usepackage{fdsymbol}

\title{Unified Pre-training for Program Understanding and Generation}

\author{
Wasi Uddin Ahmad$^\S$\thanks{~~Equal contribution.}~, Saikat Chakraborty$^\dagger$\footnotemark[1]~, Baishakhi Ray$^\dagger$, Kai-Wei Chang$^\S$ \\
$^\S$University of California, Los Angeles, $^\dagger$Columbia University \\
$^\S${\{wasiahmad, kwchang\}@cs.ucla.edu}, $^\dagger${\{saikatc, rayb\}@cs.columbia.edu}
}

\date{}
\lstdefinestyle{CustomPy}{
  belowcaptionskip=1\baselineskip,
  xleftmargin=2pt,
  xrightmargin=1pt,
  language=Python,
  numbersep=5pt,
  captionpos=b,
  tabsize=3,
  showstringspaces=false,
  basicstyle=\fontsize{9}{11}\selectfont\ttfamily,
  keywordstyle=\bfseries\color{blue!40!black},
  commentstyle=\itshape\color{green},
  identifierstyle=\color{black},
  stringstyle=\color{orange},
  numbers=left,
  stepnumber=1,
  literate={\ \ }{{\ }}1
}

\lstdefinestyle{CustomJava}{
  belowcaptionskip=1\baselineskip,
  xleftmargin=2pt,
  xrightmargin=3pt,
  language=Java,
  numbersep=5pt,
  captionpos=b,
  tabsize=3,
  showstringspaces=false,
  basicstyle=\fontsize{9}{11}\selectfont\ttfamily,
  keywordstyle=\bfseries\color{purple!40!black},
  commentstyle=\itshape\color{blue},
  identifierstyle=\color{black},
  stringstyle=\color{cyan},
  numbers=left,
  stepnumber=1,
  literate={\ \ }{{\ }}1
}

\lstdefinestyle{CustomCS}{
  belowcaptionskip=1\baselineskip,
  xleftmargin=2pt,
  xrightmargin=3pt,
  language=C++,
  numbersep=5pt,
  captionpos=b,
  tabsize=3,
  showstringspaces=false,
  basicstyle=\fontsize{9}{11}\selectfont\ttfamily,
  keywordstyle=\bfseries\color{blue!40!green},
  commentstyle=\itshape\color{orange},
  identifierstyle=\color{red!40!blue},
  stringstyle=\color{red},
  numbers=left,
  stepnumber=1,
  literate={\ \ }{{\ }}1
}

\begin{document}

\setlength{\abovedisplayskip}{6pt}
\setlength{\belowdisplayskip}{6pt}

\maketitle
\begin{abstract}
Code summarization and generation empower conversion between programming language (PL) and natural language (NL), while code translation avails the migration of legacy code from one PL to another. This paper introduces PLBART, a sequence-to-sequence model capable of performing a broad spectrum of program and language understanding and generation tasks. PLBART is pre-trained on an extensive collection of Java and Python functions and associated NL text via denoising autoencoding. Experiments on code summarization in the English language, code generation, and code translation in seven programming languages show that PLBART outperforms or rivals state-of-the-art models. Moreover, experiments on discriminative tasks, \eg program repair, clone detection, and vulnerable code detection, demonstrate PLBART's effectiveness in program understanding. Furthermore, analysis reveals that PLBART learns program syntax, style (\eg identifier naming convention), logical flow (\eg {\tt if} block inside an {\tt else} block is equivalent to {\tt else if} block) that are crucial to program semantics and thus excels even with limited annotations.
\end{abstract}

\input{latex/introduction}

\input{latex/method}
\input{latex/setup}

\input{latex/experiment}

\input{latex/relwork}

\input{latex/conclusion}



\section*{Broader Impact}
Automation in software engineering is paramount in increasing programmers' productivity. 
A reduced workload of tedious works at the part of developers' daily routine would give them more time to solve significant problems for society's wellbeing. 
There are numerous program-and-language applications in the software development lifecycle, such as code documentation/summarization, code synthesis, translating code across languages, etc that can be automated to facilitate software engineering.
The availability of large-scale data (thanks to open source repositories, forums, and millions of contributors worldwide) opens up the opportunity to solve many of those problems in a data-driven fashion. 
PLBART aims at program-and-language applications that demand a complete syntactic and semantic understanding of source code and associated textual data.  
For the tasks we have shown evaluation, PLBART will serve as a solid and replicable baseline to guide future research. 
We also believe our work could be an excellent starting point for future works aim at solving a variety of software engineering problems.

\section*{Acknowledgments}
We thank anonymous reviewers for their helpful feedback.
We also thank UCLA-NLP group for helpful discussions and comments.
This work was supported in part by National Science Foundation Grant OAC 1920462, CCF 1845893, CCF 1822965, CNS 1842456. 
Any opinions, findings, conclusions, or recommendations expressed herein are those of the authors, and do not necessarily reflect those of the US Government or NSF.

\bibliography{naacl2021,anthology}
\bibliographystyle{acl_natbib}

\clearpage
\appendix
\input{appendix/h-param}
\input{appendix/appendeix1}
\input{appendix/appendix2-code-translation}

\end{document}

%% file: latex/introduction.tex
\section{Introduction}


Engineers and developers write software programs in a programming language (PL) like Java, Python, etc., and often use natural language (NL) to communicate with each other.
Use of NL in software engineering ranges from writing documentation, commit messages, bug reports to seeking help in different forums (\eg Stack Overflow), etc. 
Automating different software engineering applications, such as source code summarization, generation, and translation, heavily rely on the understanding of PL and NL---we collectively refer them as PLUG (stands for, Program and Language Understanding and Generation) applications or tasks.
Note that the use of NL in software development is quite different than colloquially written and spoken language. 
For example, NL in software development often contains domain-specific jargon, \eg when software engineers use \emph{Code Smell\footnote{https://en.wikipedia.org/wiki/Code\_smell}}, it means a potential problem in code (something other than \emph{Smell} in regular English language).

\input{images/example}

In this work, our goal is to develop a general-purpose model that can be used in various PLUG applications.
Recent advancements in deep learning and the availability of large-scale PL and developers' NL data ushered in the automation of PLUG applications. 
One important aspect of PLUG applications is that they demand a profound understanding of program syntax and semantics and mutual dependencies between PL and NL. 
For example,~\Cref{main_example_2} shows two implementations of the same algorithm (sorting) in two PL and corresponding NL summary.
An automatic translation tool must understand that function {\tt sorted} in {Python} acts similar to {\tt Arrays.sort} in {Java} and the {\tt lambda} operation in {Python} is equivalent to instantiating a {\tt Comparator} object in {Java}.
Similarly, a tool that summarizes either of these code must understand that {\tt x[0]} in {Python} or {\tt Tuple.get(0)} in {Java} refers to the first element in the tuple list.
Most of the available data in PL and NL are unlabeled and cannot be trivially used to acquire PLUG task-specific supervision.
However, PLUG tasks have a common prerequisite \textemdash~understanding PL and NL syntax and semantics.
Leveraging unlabelled data to pretrain a model to learn PL and NL representation can be transferred across PLUG tasks.
This approach reduces the requirement of having large-scale annotations for task-specific fine-tuning.
In recent years we have seen a colossal effort to pretrain models on a massive amount of unlabeled data (\eg text, images, videos) \cite{devlin-etal-2019-bert, liu2019roberta, NIPS2019_8928, conneau-etal-2020-unsupervised, li2019visualbert, sun2019videobert}
to transfer representation encoders across a wide variety of applications.
There are a few research effort in learning general purpose PL-NL representation encoders, such as CodeBERT~\cite{feng2020codebert} and GraphCodeBERT~\cite{guo2020graphcodebert} that are pretrained on a \emph{small-scale} bimodal data (code-text pairs). 
Such models have been found effective for PLUG tasks, including code search, code completion, etc. 

Language generation tasks such as code summarization is modeled as sequence-to-sequence learning, where an encoder learns to encode the input code and a decoder generates the target summary. Despite the effectiveness of existing methods, they do not have a pretrained decoder for language generation. Therefore, they still require a large amount of parallel data to train the decoder. To overcome this limitation, \citet{lewis-etal-2020-bart} proposed denoising sequence-to-sequence pre-training where a Transformer \cite{vaswani2017attention} learns to reconstruct an original text that is corrupted using an arbitrary noise function. Very recently, \citet{lachaux2020unsupervised} studied denoising pre-training using a large-scale source code collection aiming at unsupervised program translation and found the approach useful.
This raises a natural question, \emph{can we unify pre-training for programming and natural language?}
Presumably, to facilitate such pre-training, we need unlabeled NL text that is relevant to software development.
Note that unlike other bimodal scenarios (\eg vision and language), PL and associated NL text share the same alphabet or uses anchor tokens (\eg ``sort'', ``list'', ``tuple'' as shown in \Cref{main_example_2}) that can help to learn alignment between semantic spaces across languages.

We introduce PLBART (Program and Language BART), a bidirectional and autoregressive transformer pre-trained on unlabeled data across PL and NL to learn multilingual representations applicable to a broad spectrum of PLUG applications.
We evaluate PLBART on code summarization, generation, translation, program repair, clone detection, and vulnerability detection tasks.
Experiment results show that PLBART outperforms or rivals state-of-the-art methods, \eg CodeBERT and GraphCodeBERT, demonstrating its promise on program understanding and generation.
We perform a thorough analysis to demonstrate that PLBART learns program syntax, logical data flow that is indispensable to program semantics, and excels even when limited annotations are available.
We release our code\footnote{https://github.com/wasiahmad/PLBART} to foster future research.

%% file: images/example.tex
\begin{figure}[t]
\centering
\begin{subfigure}{0.50\textwidth}
\caption{\underline{Program snippet in Python}}
\vspace{-1mm}
\begin{tabular}{l}

\lstset{escapechar=@,style=CustomPy}
\begin{lstlisting}
def sort_list(uns):
	return sorted(uns, key=lambda x:x[0])
\end{lstlisting}
\end{tabular}
\vspace{4mm}
\label{tab:python_code}
    \end{subfigure}%
\\
\vspace{-1mm}
\begin{subfigure}{0.50\textwidth}
\lstset{escapechar=@,style=CustomJava}
\caption{\underline{Program snippet in Java}}
\vspace{-1mm}
\begin{tabular}{l}
\lstset{escapechar=@,style=CustomJava}
\begin{lstlisting}
static Tuple[] sortArray(Tuple[] uns){
	return Arrays.sort(
          uns, new Comparator<Tuple>() {
		public int compare(
		 Tuple o1, Tuple o2) {
			 return o1.get(0) == o2.get(0);
		 }		
	});
}
\end{lstlisting}
\end{tabular}
\\[6pt]
\textbf{Summary:} sort a list of tuples by first element
\label{tab:java_code}
\end{subfigure}
\caption{
Example motivating the need to understand the association of program and natural languages for code summarization, generation, and translation.
}
\label{main_example_2}
\end{figure}

%% file: latex/method.tex
\section{PLBART}
\label{sec:method}
PLBART uses denoising sequence-to-sequence pre-training to utilize unlabeled data in PL and NL. Such pre-training lets PLBART reason about language syntax and semantics.
At the same time, PLBART learns to generate language coherently.

\input{table/pretrain_data}

\input{table/example1}

\subsection{Denoising Pre-training}

\paragraph{Data \& pre-processing}
We pre-train PLBART on a large-collection of Java and Python functions and natural language descriptions from Github and StackOverflow, respectively.
We download all the GitHub repositories associated with Java and Python languages available on Google BigQuery.\footnote{https://console.cloud.google.com/ marketplace/details/github/github-repos}
We extract the Java and Python functions following the pre-processing pipeline from \citet{lachaux2020unsupervised}.
We collect the StackOverflow posts (include both questions and answers, exclude code snippets) by downloading the data dump (date: 7th September 2020) from stackexchange.\footnote{https://archive.org/download/stackexchange}
Statistics of the pre-training dataset are presented in Table \ref{table:pretrain_data}.
We tokenize all the data with a sentencepiece model \cite{kudo-richardson-2018-sentencepiece} learned on 1/5'th of the pre-training data.
We train sentencepiece to learn 50,000 subword tokens.

One key challenge to aggregate data from different modalities is that some modalities may have more data, such as we have 14 times more data in PL than NL. 
Therefore, we mix and up/down sample the data following \citet{NIPS2019_8928} to alleviate the bias towards PL. 
We sample instances for pre-training according to a multinomial distribution with probabilities $(q_1, q_2, \ldots, q_N)$:
\begin{equation*}
    q_i = \frac{1}{p_i} \cdot \frac{p_i^{\alpha}}{\sum_{j=1}^N p_j^{\alpha}}, 
    p_i = \frac{n_i}{\sum_{j=1}^N n_j},
\end{equation*}
where $N$ is the total number of languages and $n_i$ is the total number of instances in language $i$.
We set the smoothing parameter $\alpha$ to 0.3.

\input{table/example2}

\paragraph{Architecture}
PLBART uses the same architecture as BART$_{base}$ \cite{lewis-etal-2020-bart}, it uses the sequence-to-sequence Transformer architecture \cite{vaswani2017attention}, with 6 layers of encoder and 6 layers of decoder with model dimension of 768 and 12 heads ($\sim$140M parameters). 
The only exception is, we include an additional layer-normalization layer on top of both the encoder and decoder following \citet{liu2020multilingual}, which is found to stabilize training with FP16 precision.

\paragraph{Noise function, $f$} 
In denoising autoencoding, a model learns to reconstruct an input text that is corrupted by a noise function.
Reconstruction of the original input requires the model to learn language syntax and semantics.
In this work, we use three noising strategies: token masking, token deletion, and token infilling \cite{lewis-etal-2020-bart}.
According to the first two strategies, random tokens are sampled and replaced with a mask token or deleted from the input sequence.
In token infilling, a number of text spans are sampled and replaced with a \emph{single} mask token.
The span lengths are drawn from a Poisson distribution ($\lambda = 3.5$).
We mask 35\% of the tokens in each instance.

\paragraph{Input/Output Format}
The input to the encoder is a noisy text sequence, while the input to the decoder is the original text with one position offset. 
A language id symbol (e.g., <java>, <python>) is appended and prepended to the encoder and decoder inputs, respectively.
We provide a few examples in Table \ref{table:pre_training}.
The input instances are truncated if they exceed a maximum sequence length of 512.

\paragraph{Learning}
PLBART is pre-trained on $N$ languages (in our case, $N$=3), where each language $N_i$ has a collection of unlabeled instances $\mathcal{D}_i = \{x_1, \ldots, x_{n_i}\}$.
Each instance is corrupted using the noise function $f$ and we train PLBART to predict the original instance $x$ from $f(x)$.
Formally, PLBART is trained to maximize $\mathcal{L}_{\theta}$:
\begin{equation*}
    \mathcal{L}_{\theta} = \sum_{i=1}^N \sum_{j=1}^{m_i} \log P (x_j | f(x_j); \theta)
\end{equation*}
where $m_i$ is the number of sampled instances in language $i$ and the likelihood $P$ is estimated following the standard sequence-to-sequence decoding.

\paragraph{Optimization} 
We train PLBART on 8 Nvidia GeForce RTX 2080 Ti GPUs for 100K steps.
The effective batch size is maintained at 2048 instances.
We use Adam ($\epsilon$ = 1e-6, $\beta_2$ = 0.98) with a linear learning rate decay schedule for optimization. 
We started the training with dropout 0.1 and reduced it to 0.05 at 50K steps and 0 at 80K steps.
This is done to help the model better fit the data \cite{liu2020multilingual}.
The total training time was approximately 276 hours (11.5 days).
All experiments are done using the Fairseq library~\cite{ott-etal-2019-fairseq}.

\subsection{Fine-tuning PLBART}
We fine-tune PLBART for two broad categories of downstream applications.

\paragraph{Sequence Generation}
PLBART has an encoder-decoder architecture where the decoder is capable of generating target sequences autoregressively.
Therefore, we can directly fine-tune PLBART on sequence generation tasks, such as code summarization, generation, and translation.
Unlike denoising pre-training, the source sequence is given as input to the encoder during fine-tuning, and the decoder generates the target sequence.
The source and target sequence can be a piece of code or text sequence.
Table \ref{table:fine_tuning} shows a few examples of input and output to and for PLBART for different generation tasks.
Note that PLBART prepends a language id to the decoded sequence; it enables fine-tuning PLBART in a multilingual setting (\eg code generation in multiple languages).\footnote{We do not perform multilingual fine-tuning in this work.}

\paragraph{Sequence Classification}
We fine-tune PLBART on sequence classification tasks following \citet{lewis-etal-2020-bart}.
The input sequence is fed into both the encoder and decoder.
For a pair of inputs, we concatenate them but insert a special token (``</s>'') between them. 
A special token is added at the end of the input sequence.
This last token's representation from the final decoder layer is fed into a linear classifier for prediction.

\paragraph{Optimization}
We fine-tune PLBART for a maximum of 100K steps on all the downstream tasks with 2500 warm-up steps.
We set the maximum learning rate, effective batch size, and dropout rate to 3e-5, 32 and 0.1, respectively.
The final models are selected based on the validation BLEU (in generation task) or accuracy (in classification tasks). 
Fine-tuning PLBART is carried out in one Nvidia GeForce RTX 2080 Ti GPU.

%% file: table/pretrain_data.tex
\begin{table}[t]
\centering
\resizebox{\linewidth}{!}{%
\small
\begin{tabular}{@{\hskip 0.05in}l@{\hskip 0.05in} r@{\hskip 0.1in} r@{\hskip 0.1in} r@{\hskip 0.05in}}
\hline
 & Java & Python & NL \\ 
\hline
All Size & 352 GB & 224 GB & 79 GB \\
All - Nb of tokens & 36.4 B & 28 B & 6.7 B \\
All - Nb of documents & 470 M & 210 M & 47 M \\
\hline
\end{tabular}
}
\caption{
Statistics of the data used to pre-train PLBART. ``Nb of documents'' refers to the number of functions in Java and Python collected from Github and the number of posts (questions and answers) in the natural language (English) from StackOverflow.
}
\label{table:pretrain_data}
\end{table}

%% file: table/example1.tex
\begin{table*}[!ht]
\centering
\resizebox{\linewidth}{!}{
\def\arraystretch{1.25}%
\begin{tabular}{l | l}
\hline
PLBART Encoder Input & PLBART Decoder Output 
\\
\hline
Is 0 the \textbf{[MASK]} Fibonacci \textbf{[MASK]} ? {\color{red} <En>} & {\color{red} <En>} Is 0 the \textbf{first} Fibonacci \textbf{number} ? \\ \hdashline
\multirow{3}{*}{\parbox{8cm}{ public static main ( String args [ ] ) \{ date = Date ( ) ; System . out . ( String . format ( " Current Date : \% tc " , ) ) ; \} {\color{blue} <java>} }} &  \multirow{3}{*}{\parbox{8cm}{{\color{blue} <java>} public static \textbf{void} main ( String args [ ] ) \{ \textbf{Date} date = \textbf{new} Date ( ) ; System . out . \textbf{printf} ( String . format ( " Current Date : \% tc " , \textbf{date} ) ) ; \} }} \\
& \\
& \\ \hdashline
\multirow{2}{*}{\parbox{8cm}{def addThreeNumbers ( x , y , z ) : NEW\_LINE INDENT return \textbf{[MASK]} {\color{orange} <python>}}} & 
\multirow{2}{*}{\parbox{8cm}{{\color{orange} <python>} def addThreeNumbers ( x , y , z ) : NEW\_LINE INDENT return \textbf{x + y + z} }} \\
& \\ 
\hline
\end{tabular}
}
\caption{
Example encoder inputs and decoder outputs during denoising pre-training of PLBART. We use three noising strategies: token masking, token deletion, and token infilling (shown in the three examples, respectively).
}
\label{table:pre_training}
\end{table*}

%% file: table/example2.tex
\begin{table*}[!ht]
\centering
\resizebox{\linewidth}{!}{
\def\arraystretch{1.1}%
\begin{tabular}{l|l|l}
\hline
& \tool Encoder Input & \tool Decoder Input \\
\hline
\multirow{2}{*}{S} & \multirow{2}{*}{\parbox{7.5cm}{
def maximum (a , b , c) : NEW\_LINE INDENT return max ( [ a , b , c ] ) {\color{orange} <python>}}} &  \multirow{2}{*}{\parbox{7.5cm}{{\color{red} <En>} Find the maximum of three numbers}} \\
& & \\ \hdashline
\multirow{2}{*}{G} & \multirow{2}{*}{\parbox{7.5cm}{Find the maximum of three numbers {\color{red} <En>}}} & \multirow{2}{*}{\parbox{7.5cm} { {\color{blue} <java>} public int maximum ( int a , int b , int c ) \{ return Math . max ( a , Math . max ( b , c ) ) \}}} \\
& & \\ \hdashline
\multirow{2}{*}{T} & \multirow{2}{*}{\parbox{7.5cm}{ public int maximum ( int a , int b , int c ) \{ return Math . max ( a , Math . max ( b , c ) ) \} {\color{blue} <java>} }} & \multirow{2}{*}{\parbox{7.5cm}{{\color{orange} <python>} def maximum (a , b , c) : NEW\_LINE INDENT return max ( [ a , b , c ] )}} \\
& & \\
\hline
\end{tabular}
}
\caption{Example inputs to the encoder and decoder for fine-tuning PLBART on sequence generation tasks: source code summarization (S), generation (G), and translation (T).}
\label{table:fine_tuning}
\end{table*}

%% file: latex/setup.tex
\section{Experiment Setup}
\label{sec:exp_setup}

\input{table/github_datastat}


To understand PLBART's performance in a broader context, we evaluate PLBART on several tasks. 
Our evaluation focuses on assessing PLBART's ability to capture rich semantics in source code and associated natural language text.

\subsection{Evaluation Tasks}
We divide the evaluation tasks into four categories.
The evaluation task datasets are summarized in Table \ref{tab:dataset_task}.
We use CodeXGLUE~\cite{CodeXGLUE} provided public dataset and corresponding train-validation-test splits for all the tasks.

\paragraph{Code Summarization}
refers to the task of generating a natural language (English) summary from a piece of code. We fine-tune PLBART on summarizing source code written in six different programming languages, namely, Ruby, Javascript, Go, Python, Java, and PHP.

\paragraph{Code Generation} 
is exactly the opposite of code summarization. It refers to the task of generating a code (in a target PL) from its NL description.
We fine-tune PLBART on the Concode dataset~\cite{iyer-etal-2018-mapping}, where the input is a text describing class member functions in Java and class environment, the output is the target function.

\paragraph{Code Translation}
requires a model to generate an equivalent code in the target PL from the input code written in the source PL.
Note that the source and target PL can be the same.
Hence, we consider two types of tasks in this category.

The first task is a typical PL translation task, translating a code \ie from Java code to C\#, and vice versa.
In this task, the semantic meaning of the translated code should exactly match the input code. 
Thus, this task evaluates PLBART's understanding of program semantics and syntax across PL.
The second task we consider is program repair. 
In this task, the input is a buggy code, and the output is a modified version of the same code which fixes the bug. 
This task helps us understand PLBART's ability to understand code semantics and apply semantic changes in the code. 

\paragraph{Code Classification}
aims at predicting the target label given a single or a pair of source code. 
We evaluate PLBART on two classification tasks. 
The first task is clone detection, where given a pair of code, the goal is to determine whether they are clone of each other (similar to paraphrasing in NLP).
The second task is detecting whether a piece of code is vulnerable.
This task help us gauging PLBART's effectiveness in program understanding in an unseen PL since the code examples in this task are written in C/C++.

\subsection{Evaluation Metrics}

\paragraph{BLEU} 
computes the n-gram overlap between a generated sequence and a collection of references.
We use corpus level BLEU~\cite{papineni-etal-2002-bleu} score for all the generation tasks, except code summarization where we use smoothed BLEU-4 score \cite{lin-och-2004-orange} following \citet{feng2020codebert}. 


\paragraph{CodeBLEU}
is a metric for measuring the quality of the synthesized code~\cite{ren2020codebleu}. 
Unlike BLEU, CodeBLEU also considers grammatical and logical correctness based on the abstract syntax tree and the data-flow structure. 

\paragraph{Exact Match (EM)}
evaluates if a generated sequence exactly matches the reference. 


\subsection{Baseline Methods}
We compare PLBART with several state-of-the-art models and broadly divide them into two categories.
First, the models that are trained on the evaluation tasks from scratch, and second, the models that are pre-trained on unlabeled corpora and then fine-tuned on the evaluation tasks.

\subsubsection{Training from Scratch}


\noindent{\textbf{Seq2Seq}~\cite{luong-etal-2015-effective}\hspace{0.5em}} 
is an LSTM based Seq2Seq model with attention mechanism. Vocabulary is constructed using byte-pair encoding.

\medskip
\noindent{\textbf{Transformer}~\cite{vaswani2017attention}\hspace{0.5em}}
is the base architecture of PLBART and other pre-trained models.
Transformer baseline has the same number of parameters as PLBART.
Hence, a comparison with this baseline demonstrates the direct usefulness of pre-training PLBART.

\subsubsection{Pre-trained Models}
As described in section~\ref{sec:method}, PLBART consists of an encoder and autoregressive decoder. 
We compare PLBART on two categories of pre-trained models. First, the encoder-only models (\eg RoBERTa, CodeBERT, and GraphCodeBERT) that are combined with a randomly initialized decoder for task-specific fine-tuning. The second category of baselines include decoder-only models (CodeGPT) that can perform generation autoregressively.

\input{table/code_to_text}

\medskip
\noindent\textbf{RoBERTa, RoBERTa (code) \hspace{0.5em}} 
are RoBERTa \cite{liu2019roberta} model variants.
While RoBERTa is pre-trained on natural language, RoBERTa (code) is pre-trained on source code from CodeSearchNet~\cite{husain2019codesearchnet}.

\medskip
\noindent{\textbf{CodeBERT}~\cite{feng2020codebert} \hspace{0.5em}} 
combines masked language modeling (MLM)~\cite{devlin-etal-2019-bert} with replaced token detection objective~\cite{clark2020electra} to pretrain a Transformer encoder.

\medskip
\noindent{\textbf{GraphCodeBERT} \cite{guo2020graphcodebert} \hspace{0.5em}}
is a concurrent work with this research which improved CodeBERT by modeling the data flow edges between code tokens.
We report GraphCodeBERT's performance directly from the paper since their implementation is not publicly available yet.

\medskip
\noindent\textbf{GPT-2, CodeGPT-2, and CodeGPT-adapted \hspace{0.5em}}
are GPT-style models.
While GPT-2 \cite{radford2019language} is pretrained on NL corpora, CodeGPT-2 and CodeGPT-adapted are pretrained on CodeSearchNet \cite{CodeXGLUE}.
Note that, CodeGPT-adapted starts from the GPT-2 checkpoint for pre-training.



%% file: table/github_datastat.tex
\begin{table*}[!ht]
\centering
\begin{tabular}{l|l|l|r|r|r}
\hline
Task & {Dataset} & Language & Train & Valid & Test \\
\hline
\multirow{6}{*}{Summarizaion} & \multirow{6}{*}{\citet{husain2019codesearchnet}} &Ruby & 24,927& 1,400& 1,261\\
& & Javascript & 58,025	& 3,885 &	3,291 \\
& & Go & 167,288	&   7,325	&   8,122\\
& & Python & 251,820	&   13,914	&   14,918 \\
& & Java & 164,923	&   5,183	&   10,955 \\
& & PHP & 241,241	&   12,982	&   14,014\\
\hline
Generation  & \citet{iyer-etal-2018-mapping} & NL to Java & 100,000 & 2,000 & 2,000\\
\hline
\multirow{4}{*}{Translation} & \multirow{2}{*}{Code-Code \cite{CodeXGLUE}}  & Java to C\# & 10,300 & 500 & 1,000\\
& & C\# to Java & 10,300 & 500 & 1,000\\
\cline{2-6}
& Program Repair & Java$_{small}$ & 46,680 & 5,835 & 5,835\\
& \cite{tufano2019empirical}& Java$_{medium}$ & 52,364 & 6,545 &6,545\\
\hline
\multirow{4}{*}{Classification}& Vulnerability Detection & \multirow{2}{*}{C/C++} & \multirow{2}{*}{21,854} & \multirow{2}{*}{2,732} & \multirow{2}{*}{2,732} \\
& \cite{zhou2019devign} & & & & \\
\cline{2-6}
& Clone Detection & \multirow{2}{*}{Java} & \multirow{2}{*}{100,000} & \multirow{2}{*}{10,000} & \multirow{2}{*}{415,416} \\
& \cite{wang2020detecting} & & & & \\
\hline
\end{tabular}
\caption{Statistics of the downstream benchmark datasets.}
\label{tab:dataset_task}
\end{table*}

%% file: table/code_to_text.tex
\begin{table*}[!ht]
\centering
\begin{tabular}{l|c c c c c c|c}
\hline
Methods & Ruby & Javascript & Go & Python & Java & PHP & Overall \\
\hline
Seq2Seq &  9.64	& 10.21	& 13.98	& 15.93	& 15.09	& 21.08	& 14.32 \\
Transformer & 11.18	& 11.59	& 16.38	& 15.81	& 16.26	& 22.12 & 15.56 \\
RoBERTa  & 11.17 & 11.90 & 17.72 & 18.14 & 16.47 & 24.02 & 16.57 \\
CodeBERT & 12.16 & 14.90 & 18.07 & 19.06 & 17.65 & {\bf 25.16} & 17.83 \\ \hline
\tool & {\bf 14.11} & {\bf 15.56} & {\bf 18.91} & {\bf 19.30} & {\bf 18.45} & 23.58 & {\bf 18.32} \\
\hline
\end{tabular}
\caption{Results on source code summarization, evaluated with smoothed BLEU-4 score. The baseline results are reported from \citet{feng2020codebert}.}
\label{table:code_to_text}
\end{table*}

%% file: latex/experiment.tex
\section{Results \& Analysis}
We aim to address the following questions.
\begin{compactenum}
    \item Does PLBART learn strong program and language representations from unlabeled data?
    \item Does PLBART learn program characteristics, \eg syntax, style, and logical data flow?
    \item How does PLBART perform in an unseen language with limited annotations?
\end{compactenum}

\input{table/text_to_code}
\input{images/qual_example_concode}
\input{table/code_to_code}

\subsection{Code Summarization}

\Cref{table:code_to_text} shows the result of code summarization.
PLBART outperforms the baseline methods in five out of the six programming languages with an overall average improvement of 0.49 BLEU-4 over CodeBERT.
The highest improvement ($\sim$16\%) is in the Ruby language, which has the smallest amount of training examples.
Unlike CodeBERT, PLBART is not pretrained on the Ruby language; however, the significant performance improvement indicates that PLBART learns better generic program semantics.
In contrast, PLBART performs poorly in the PHP language. The potential reason is syntax mismatch between the pre-trained languages and PHP. 
Surprisingly, RoBERTa performs better than PLBART on the PHP language. 
We suspect that since RoBERTa is pre-trained on natural language only, it does not suffer from the syntax mismatch issue.
Overall in comparison to the Transformer baseline, PLBART improves with an average of 2.76 BLEU-4, and we credit this improvement to the pre-training step.


\subsection{Code Generation}

\Cref{table:concode} shows the evaluation result on code generation from NL description.
PLBART outperforms all the baselines in terms of BLEU and CodeBLEU. 
While CodeGPT-adapted~\cite{CodeXGLUE} achieves the best Exact Match (EM) score, PLBART outperforms CodeGPT-adapted by a large margin in terms of CodeBLEU. 
This result implies that PLBART generates \emph{significantly more} syntactically and logically correct code than all the baselines.

~\Cref{concode_example} shows an example of code generated by PLBART.
The difference between the reference code and the generated code is in line 6 onward. In the reference code, {\tt loc0} is returned, however same {\tt loc0} is returned in an {\tt else} block in the generated code. If we look closely, in the reference code, line 6 will be executed only if the condition in line 3 (\ie {\tt loc0 == null}) is {\tt false}. In the generated code, {\tt loc0} will be returned only if the condition in line 3 is {\tt false}, making the generated code semantically equivalent to the reference code. 

To study whether PLBART learns code syntax and logical flow during pre-training or fine-tuning, we perform an ablation study where we use subset of the training examples (10K, 20K, and 50K) to fintune PLBART in this task.
As \cref{table:concode} shows, with only 10K examples, PLBART outperforms all baselines in terms of CodeBLUE. 
This ablation shows that PLBART learns program syntax and data flow during pre-training, resulting in effective performance on downstream tasks even when finetuned on small number of examples.

As shown in prior works~\cite{yin-neubig-2017-syntactic, chakraborty2020codit}, generating syntactically and logically correct code has been a big challenge in program generation.
We conjecture that PLBART's large-scale denoising sequence-to-sequence pre-training helps understand program syntax and logical flow; therefore enables PLBART to generate syntactically and logically valid code.

\subsection{Code Translation}

\Cref{table:code_translation} presents the evaluation results on code translation.
PLBART outperforms all the baselines \wrt EM, BLEU, and CodeBLEU. 
PLBART improves over CodeBERT by 9.5\% and 10.5\% when translating from Java to C\# and C\# to Java, respectively.
Although PLBART is not pretrained on C\# language, there is a significant syntactic and semantic similarity between Java and C\#. Thus PLBART understands C\# language syntax and semantics. 
However, such similarities are non-trivial, making the Naive copy and PBSMT perform very poorly in both the translation tasks.

\input{images/qual_example_cs_to_java}

\Cref{code_trans_example} shows an example where PLBART's generated C\# code does not exactly match the reference; however, they are semantically equivalent. 
In the reference, the {\tt else} block (line 4-9) is equivalent to the {\tt else if} block (line 4-7) in the generated code.
In addition, {\tt start} is generated as function parameter and used in the function body, equivalent to {\tt start\_1} in the reference code. 
This further corroborates the syntactic understanding of PLBART and its ability to reason about the data flow in source code. 
We present more qualitative examples in Appendix.

In the program repair task, both the input and the output are in the same language.
While the input is a buggy code, the output should be the target bug-free code.
Thus in this task, the exact match is the critical metric.
Nevertheless, as shown in~\cref{table:bug_fix}, PLBART can generate 17.13\%, and 74.03\% more correct bug fixes than CodeBERT in Java$_{small}$ and Java$_{medium}$ datasets, respectively. 
On the other hand, PLBART performs comparably to GraphCodeBERT that uses structure-aware pre-training to learn program syntax and semantics.


\subsection{Classification}

In both clone detection and the vulnerability detection tasks, PLBART outperforms CodeBERT. We present the results in Table \ref{table:classification}.
In the vulnerability detection task, code semantics is the most critical feature \cite{zhou2019devign, chakraborty2020deep}. 
Since PLBART is not pretrained on C/C++ language, its improved performance compared to the Transformer baseline is the testament that PLBART can identify semantics beyond the language syntax's specifics.
Moreover, PLBART's improved performances over CodeBERT and GraphCodeBERT confirms its effectiveness in program understanding in addition to its generation ability.

\input{table/repair}

\input{table/vulnerability}

We acknowledge that neither PLBART nor CodeBERT is state-of-the-art in vulnerability detection, as graph-based models perform best in this task. 
In this evaluation, our goal is to study how well PLBART understands program semantics in an unseen language for a different type of task (other than the generation, \ie classification).

%% file: table/text_to_code.tex
\begin{table}[t]
\centering
\resizebox{\linewidth}{!}{%
\begin{tabular}{l|c c c}
\hline
Methods & EM & BLEU & CodeBLEU \\ 
\hline
Seq2Seq & 3.05 & 21.31 & 26.39 \\
\citet{guo-etal-2019-coupling} & 10.05 & 24.40 & 29.46 \\
\citet{iyer-etal-2019-learning} & 12.20 & 26.60 & - \\
GPT-2 & 17.35 & 25.37 & 29.69 \\
CodeGPT-2 & 18.25 & 28.69 & 32.71 \\
CodeGPT-adapted & {\bf 20.10} & 32.79 & 35.98 \\ \hline
PLBART & 18.75 & {\bf 36.69} & {\bf 38.52} \\
\hline
\hline
PLBART$_{10K}$ & 17.25 & 31.40 & 33.32 \\
PLBART$_{20K}$ & 18.45 & 34.00 & 35.75 \\
PLBART$_{50K}$ & 17.70 & 35.02 & 37.11 \\
\hline
\end{tabular}
}
\caption{Results on text-to-code generation task using the CONCODE dataset \cite{iyer-etal-2018-mapping}.}
\label{table:concode}
\end{table}

%% file: images/qual_example_concode.tex
\begin{figure}[t]
\rule{\linewidth}{0.6pt}
{{\bf Input text:} returns the count to which the specified key is mapped in this frequency counter , or 0 if the map contains no mapping for this key .}
\\ [8pt]
\begin{subfigure}{0.50\textwidth}
\caption{\underline{Reference Code}}
\vspace{-1mm}
\begin{tabular}{l}
\lstset{escapechar=@,style=CustomJava}
\begin{lstlisting}
Integer function (T arg0) { 
	Integer loc0 = counter.get(arg0); 
	if (loc0 == null) { 
		return 0 ;
	} 
	return loc0; 
}
\end{lstlisting}
\end{tabular}
\label{tab:ref_code_concode}
\end{subfigure}%
\\ \\
\begin{subfigure}{0.50\textwidth}
\lstset{escapechar=@,style=CustomJava}
\caption{\underline{Generated Code}}
\vspace{-1mm}
\begin{tabular}{l}
\lstset{escapechar=@,style=CustomJava}
\begin{lstlisting}
int function (T arg0) { 
	Integer loc0 = counter.get(arg0); 
	if (loc0 == null) { 
		return 0 ;
	} 
	else { 
		return loc0;
	} 
}
\end{lstlisting}
\end{tabular}
\label{tab:gen_code_concode}
\end{subfigure}
\rule{\linewidth}{0.6pt}
\caption{
An example of generated code by PLBART that is syntactically and semantically valid, but does not match the reference.
}
\label{concode_example}
\vspace{-2mm}
\end{figure}

%% file: table/code_to_code.tex
\begin{table*}[!ht]
\centering
\begin{tabular}{l|c c c| c c c}
\hline
\multirow{ 2}{*}{Methods} & \multicolumn{3}{c|}{Java to C\#} & \multicolumn{3}{c}{C\# to Java} \\ 
\cline{2-7}
& BLEU & EM & CodeBLEU & BLEU & EM & CodeBLEU \\ 
\hline
Naive Copy & 18.54 & 0 & 34.20 & 18.69 & 0 & 43.04 \\
PBSMT & 43.53 & 12.50 & 42.71 & 40.06 & 16.10 & 43.48 \\
Transformer & 55.84	& 33.00	& 63.74	& 50.47	& 37.90	& 61.59 \\
RoBERTa (code)  & 77.46 & 56.10 & 83.07 & 71.99 & 57.90 & 80.18 \\
CodeBERT & 79.92 & 59.00 & 85.10 & 72.14 & 58.80 & 79.41 \\ 
GraphCodeBERT & 80.58 & 59.40 & - & 72.64 & 58.80 & - \\\hline

\tool  & {\bf 83.02} & {\bf 64.60} & {\bf 87.92} & {\bf 78.35} & {\bf 65.00} & {\bf 85.27} \\
\hline
\end{tabular}
\caption{
Results on source code translation using Java and C\# language dataset introduced in \cite{CodeXGLUE}.
PBSMT refers to phrase-based statistical machine translation where the default settings of Moses decoder \cite{koehn-etal-2007-moses} is used. The training data is tokenized using the RoBERTa \cite{liu2019roberta} tokenizer. 
}
\label{table:code_translation}
\end{table*}

%% file: images/qual_example_cs_to_java.tex
\begin{figure}[t]
\rule{\linewidth}{0.6pt}
\begin{subfigure}{0.50\textwidth}
\caption{\underline{Reference Code : C\#}}
\vspace{-1mm}
\begin{tabular}{l}
\lstset{escapechar=@,style=CustomCS}
\begin{lstlisting}
public bool find(int start_1){
	findPos = start_1;
	...
	else{
		if (findPos >= _regionEnd){
			matchFound = false;
			return false;
		}
	}
	...
}
\end{lstlisting}
\end{tabular}
\label{tab:Example_1_Cs_to_java}
\end{subfigure}%
\\ \\
\begin{subfigure}{0.50\textwidth}
\lstset{escapechar=@,style=CustomJava}
\caption{\underline{Generated Code : C\#}}
\vspace{-1mm}
\begin{tabular}{l}
\lstset{escapechar=@,style=CustomCS}
\begin{lstlisting}
public bool find(int start){
	findPos = start;
	...
	else if (findPos >= _regionEnd){
		matchFound = false;
		return false;
	}
	...
}
\end{lstlisting}
\end{tabular}
\end{subfigure}
\rule{\linewidth}{0.6pt}
\caption{
Example C\# code generated by \tool that does not exactly match the reference code.
}
\label{code_trans_example}
\vspace{-2mm}
\end{figure}

%% file: table/repair.tex
\begin{table}[!ht]
\centering
\resizebox{\linewidth}{!}{%
\begin{tabular}{l|c@{\hskip 0.1in} c|c@{\hskip 0.1in} c}
\hline
\multirow{2}{*}{Methods} & \multicolumn{2}{c|}{Small} & \multicolumn{2}{c}{Medium} \\ 
\cline{2-5}
& EM & BLEU & EM & BLEU \\ 
\hline
Naive Copy & 0 & {78.06} & 0 & {90.91} \\
Seq2Seq & 10.00 & 76.76 & 2.50	& 72.08 \\
Transformer  & 14.70 & 77.21 & 3.70 & 89.25 \\
CodeBERT & 16.40 & 77.42 & 5.16 & \textbf{91.07} \\
GraphCodeBERT & 17.30 & \textbf{80.58} & {\bf 9.10} & 72.64 \\
\hline
\tool & {\bf 19.21} & 77.02 & {8.98} & 88.50 \\
\hline
\end{tabular}
}
\caption{Results on program repair (in Java).}
\label{table:bug_fix}
\end{table}

%% file: table/vulnerability.tex
\begin{table}[t]
\centering
\begin{tabular}{l|c c}
\hline
\multirow{2}{*}{Tasks} & Vulnerability & Clone \\ 
& Detection & Detection \\
\hline
Transformer & 61.64  & - \\
CodeBERT & 62.08 & 96.5 \\
GraphCodeBERT & - & 97.1 \\ 
\hline
\tool & {\bf 63.18} & {\bf 97.2} \\
\hline
\end{tabular}
\caption{Results on the vulnerable code detection (accuracy) and clone detection (F1 score) tasks.}
\label{table:classification}
\end{table}

%% file: latex/relwork.tex
\section{Related Work}

\paragraph{Pre-training for Language Understanding and Generation}
Transformer~\cite{vaswani2017attention}, a sequence-to-sequence architecture that includes an encoder and decoder, has shown tremendous promise in natural language processing (NLP), computer vision, software engineering, and more.
\citet{devlin-etal-2019-bert} first proposed to pre-train a large Transformer architecture, called BERT, to learn representations of natural language using large-scale unlabeled data in a self-supervised fashion.
Later, BERT's task-independent pre-training approach is rigorously studied~\cite{devlin-etal-2019-bert, liu2019roberta, solaiman2019release, feng2020codebert, sun2019videobert, li-etal-2020-bert-vision}. 
While BERT-like models have shown effectiveness in learning contextualized representation, it is not very useful in generation tasks.
GPT~\cite{radford2018improving} style models improve upon BERT for generative tasks with autoregressive pre-training; however, unlike BERT, they are not bidirectional.
\citet{lewis-etal-2020-bart} introduced BART, a denoising autoencoder that uses a bidirectional encoder and an auto-regressing decoder.
Similar to BART, PLBART uses denoising pre-training to cope with generative tasks and learns multilingual representations of programming and natural language jointly. 

\medskip
\noindent\textbf{Deep Learning in Software Engineering \hspace{0.5em}}
There is a growing interest in automating software engineering (SE) using deep learning in the last few years. 
Vast sources of code in open source repositories and forums make deep learning feasible for SE tasks.
Code Summarization~\cite{movshovitz2013natural, allamanis2016convolutional, iyer2016summarizing, alon2018codeseq, hu2018deep, harer2019tree, ahmad-etal-2020-transformer}, Bug Detection~\cite{ray2016naturalness, li2018vuldeepecker, russell2018automated, zhou2019devign, chakraborty2020deep},  Program Repair~\cite{chen2019sequencer, chakraborty2020codit, lutellier2020coconut}, Code Translation~\cite{chen2018tree, drissi2018program, xu2020tree2tree}, Clone Detection~\cite{zhang2019novel, yu2019neural, wang2020detecting}, Code completion~\cite{li2017code, hellendoorn2017fse, parvez-etal-2018-building} are some of the tasks that are addressed with deep neural solution.
While most of the prior approaches use task-specific representation learning, a few works \cite{alon2019code2vec, feng2020codebert, guo2020graphcodebert, lachaux2020unsupervised, clement-etal-2020-pymt5} attempted to learn transferable representations in an unsupervised fashion.
More closely to our work, CodeBERT~\cite{feng2020codebert} is pre-trained on bimodal data to capture the semantic interaction between the input modalities (\ie program and natural languages).
More recently, GraphCodeBERT~\cite{guo2020graphcodebert} improves upon CodeBERT by leveraging data flow in source code.
In contrast, PLBART is pre-trained on large-scale data using denoising autoencoding to learn the program and natural language representations that make it effective for a broad spectrum of software engineering tasks.




%% file: latex/conclusion.tex
\section{Conclusion}
This paper presents PLBART, a sizeable pre-trained sequence-to-sequence model that can perform program and language understanding and generation tasks. PLBART achieves state-of-the-art performance on various downstream software engineering tasks, including code summarization, code generation, and code translation. Furthermore, experiments on discriminative tasks establish PLBART's effectiveness on program understanding. We also show that PLBART learns crucial program characteristics due to pre-training, such as syntax, identifier naming conventions, data flow. In the future, we want to explore ways to fine-tune PLBART on all the downstream tasks jointly.

%% file: appendix/h-param.tex
\begin{table*}[!htb]
\centering
\begin{tabular}{l|c|c|c|c|c}
\hline
Hyper-parameter & RoBERTa$^\ast$ & CodeGPT-2 & CodeBERT & GraphCodeBERT & PLBART \\ 
\hline
vocab size & 50,265 & 50,001 & 50,265 & - & 50,004 \\
n\_positions & 514 & 1024 & 514 & 256 & 1024 \\
model size & 768 & 768 & 768 & 768 & 768 \\
\# layers & 12 & 12 & 12 & 12 & 6 \\
\# heads & 12 & 12 & 12 & 12 & 12 \\
d$_{ff}$ & 3072 & 3072 & 3072 & - & 3072 \\
dropout & 0.1 & 0.1 & 0.1 & - & 0.1 \\
optimizer & Adam & Adam & Adam & Adam & Adam \\
learning rate & 5e-5 & 5e-5 & 5e-5 & 1e-4 & 5e-5 \\
batch size & 32 & 32 & 32 & 64 & 32 \\
\hline
\end{tabular}
\caption{Details of the hyper-parameters used during fine-tuning for sequence generation tasks. $^\ast$ indicates pre-trained from scratch using source code-text pairs.
}
\label{tab:h_param_details}
\end{table*}

%% file: appendix/appendeix1.tex
\begin{figure*}[!htb]
\rule{\textwidth}{1.5pt}
\vspace{-1mm}
{Example 1 : get the msg value. 
} \\
\rule{\textwidth}{0.3pt}
\\
\begin{subfigure}{0.50\textwidth}
\caption{\underline{Reference Code}}
\vspace{-1mm}
\begin{tabular}{l}
\lstset{escapechar=~,style=CustomJava}
\begin{lstlisting}
String function() { 
	return msg; 
}
\end{lstlisting}
\end{tabular}
\end{subfigure}%
\begin{subfigure}{0.50\textwidth}
\lstset{escapechar=@,style=CustomJava}
\caption{\underline{Generated Code}}
\vspace{-1mm}
\begin{tabular}{l}
\lstset{escapechar=@,style=CustomJava}
\begin{lstlisting}
String function() { 
	return this.msg; 
}
\end{lstlisting}
\end{tabular}
\end{subfigure}

\rule{\linewidth}{1.5pt}

{Example 2 : returns the instance of the singleton . 
} \\
\rule{\textwidth}{0.3pt}
\begin{subfigure}{0.50\textwidth}
\caption{\underline{Reference Code}}
\vspace{-1mm}
\begin{tabular}{l}
\lstset{escapechar=~,style=CustomJava}
\begin{lstlisting}
IConfigurationFactory function() { 
    return SINGLETON; 
}
\end{lstlisting}
\end{tabular}
\end{subfigure}%
\begin{subfigure}{0.50\textwidth}
\lstset{escapechar=~,style=CustomJava}
\caption{\underline{Generated Code}}
\vspace{-1mm}
\begin{tabular}{l}
\lstset{escapechar=@,style=CustomJava}
\begin{lstlisting}
IConfigurationFactory function() { 
    if (SINGLETON == null){ 
        SINGLETON = new SINGLETON(); 
    } 
    return SINGLETON; 
}
\end{lstlisting}
\end{tabular}
\end{subfigure}

\rule{\linewidth}{1.5pt}

{Example 3 : convert the reader into an inputstream . 
}\\
\rule{\textwidth}{0.3pt}
\begin{subfigure}{0.50\textwidth}
\lstset{escapechar=@,style=CustomJava}
\caption{\underline{Reference Code}}
\vspace{-1mm}
\begin{tabular}{l}
\lstset{escapechar=~,style=CustomJava}
\begin{lstlisting}
InputStream function () { 
    return new ReaderInputStream(reader);
}
\end{lstlisting}
\end{tabular}
\end{subfigure}%
\begin{subfigure}{0.50\textwidth}
\caption{\underline{Generated Code}}
\vspace{-1mm}
\begin{tabular}{l}
\lstset{escapechar=@,style=CustomJava}
\begin{lstlisting}
InputStream function () { 
    return reader.getInputStream(); 
}
\end{lstlisting}
\end{tabular}
\end{subfigure}

\rule{\linewidth}{1.5pt}

{Example 4 : setter for a property . if the property already exists , the value will be overridden .
}\\
\rule{\textwidth}{0.3pt}
\begin{subfigure}{0.50\textwidth}
\lstset{escapechar=@,style=CustomJava}
\caption{\underline{Reference Code}}
\vspace{-1mm}
\begin{tabular}{l}
\lstset{escapechar=~,style=CustomJava}
\begin{lstlisting}
void function ( 
	final String arg0, 
	final String arg1 ) { 
	properties.setProperty ( 
		arg0, arg1) ; 
}
\end{lstlisting}
\end{tabular}
\end{subfigure}%
\begin{subfigure}{0.50\textwidth}
\caption{\underline{Generated Code}}
\vspace{-1mm}
\begin{tabular}{l}
\lstset{escapechar=@,style=CustomJava}
\begin{lstlisting}
void function ( 
	String arg0, 
	String arg1) { 
	properties.put ( 
		arg0, arg1) ; 
}
\end{lstlisting}
\end{tabular}
\end{subfigure}

\rule{\linewidth}{1.5pt}

{Example 5 : clear the buffer . 
}\\
\rule{\textwidth}{0.3pt}
\begin{subfigure}{0.50\textwidth}
\lstset{escapechar=@,style=CustomJava}
\caption{\underline{Reference Code}}
\vspace{-1mm}
\begin{tabular}{l}
\lstset{escapechar=~,style=CustomJava}
\begin{lstlisting}
void function() {  
	bufferSize = 0; 
}
\end{lstlisting}
\end{tabular}
\end{subfigure}%
\begin{subfigure}{0.50\textwidth}
\caption{\underline{Generated Code}}
\vspace{-1mm}
\begin{tabular}{l}
\lstset{escapechar=@,style=CustomJava}
\begin{lstlisting}
void function() {  
	buffer = null; 
}
\end{lstlisting}
\end{tabular}
\end{subfigure}
\rule{\linewidth}{1.5pt}
\caption{
Qualitative examples of ``text to code'' generation using PLBART, where generated code does not exactly match the reference. 
}
\label{text_to_code_gen_examples}
\vspace{-2mm}
\end{figure*}

%% file: appendix/appendix2-code-translation.tex
\input{images/code_translation}

%% file: images/code_translation.tex
\begin{figure*}[!htb]
\rule{\textwidth}{0.6pt}
{{Example 1 : \ding{52}}} \\
\begin{subfigure}{0.50\textwidth}
\caption{\underline{Input Code : C\#}}
\begin{tabular}{l}
\lstset{escapechar=~,style=CustomCS}
\begin{lstlisting}
public int GetCells(){
	int size = 0;
	foreach (char c in cells.Keys){
		Cell e = At(c);
		if (e.cmd >= 0 || e.@ref >= 0){
			size++;
		}
	}
	return size;
}
\end{lstlisting}
\end{tabular}
\end{subfigure}%
\begin{subfigure}{0.50\textwidth}
\lstset{escapechar=@,style=CustomJava}
\caption{\underline{Generated Code : Java}}
\begin{tabular}{l}
\lstset{escapechar=@,style=CustomJava}
\begin{lstlisting}
public int getCells() {
	Iterator<Character> i = 
	    cells.keySet().iterator();
	int size = 0;
	for (; i.hasNext();) {
		Character c = i.next();
		Cell e = at(c);
		if (e.cmd >= 0 || e.ref >= 0){
			size++;
		}
	}
	return size;
}
\end{lstlisting}
\end{tabular}
\end{subfigure}
\rule{\linewidth}{0.6pt}

{{Example 2 : \ding{52}}} \\
\begin{subfigure}{0.50\textwidth}
\caption{\underline{Input Code : C\#}}
\begin{tabular}{l}
\lstset{escapechar=~,style=CustomCS}
\begin{lstlisting}
public override Cell Merge(
    Cell m, Cell e){
	if (m.cmd == e.cmd && m.@ref == 
	  e.@ref && m.skip == e.skip) {
		Cell c = new Cell(m);
		c.cnt += e.cnt;
		return c;
	} 
	else {
		return null;
	}
}
\end{lstlisting}
\end{tabular}
\end{subfigure}%
\begin{subfigure}{0.50\textwidth}
\lstset{escapechar=~,style=CustomJava}
\caption{\underline{Generated Code : Java}}
\begin{tabular}{l}
\lstset{escapechar=@,style=CustomJava}
\begin{lstlisting}
public Cell merge(Cell m, Cell e) {
	if (m.cmd == e.cmd && 
		m.ref == e.ref && 
		m.skip == e.skip) {
		Cell c = new Cell(m);
		c.cnt += e.cnt;
		return c;
	} 
	else {
		return null;
	}
}
\end{lstlisting}
\end{tabular}
\end{subfigure}

\rule{\linewidth}{0.6pt}

{{Example 3 : \ding{52}}}\\
\begin{subfigure}{0.50\textwidth}
\lstset{escapechar=@,style=CustomJava}
\caption{\underline{Input Code : Java}}
\begin{tabular}{l}
\lstset{escapechar=~,style=CustomJava}
\begin{lstlisting}
@Override public List<E> subList(
    int start, int end) {
	synchronized (mutex) {
		return new 
		    SynchronizedRandomAccessList
		    <E>(
			list.subList(start, end), 
			mutex);
	}
}
\end{lstlisting}
\end{tabular}
\end{subfigure}%
\begin{subfigure}{0.50\textwidth}
\caption{\underline{Generated Code : C\#}}
\begin{tabular}{l}
\lstset{escapechar=@,style=CustomCS}
\begin{lstlisting}
public override java.util.List<E> 
    SubList(int start, int end){
	lock (mutex){
		return new java.util.Collections.
		    SynchronizedRandomAccessList<E>(
			list.subList(start, end), 
			mutex
		);
	}
}
\end{lstlisting}
\end{tabular}
\end{subfigure}

\rule{\linewidth}{0.6pt}

{{Example 4 : \ding{52}}}\\
\begin{subfigure}{0.50\textwidth}
\lstset{escapechar=@,style=CustomJava}
\caption{\underline{Input Code : Java}}
\begin{tabular}{l}
\lstset{escapechar=~,style=CustomJava}
\begin{lstlisting}
public 
    PersianNormalizationFilterFactory(
		Map<String,String> args) {
	super(args);
	if (!args.isEmpty()) {
		throw 
		    new IllegalArgumentException(
			    "Unknown parameters: " 
			    + args);
	}
}
\end{lstlisting}
\end{tabular}
\end{subfigure}%
\begin{subfigure}{0.50\textwidth}
\caption{\underline{Generated Code : C\#}}
\begin{tabular}{l}
\lstset{escapechar=@,style=CustomCS}
\begin{lstlisting}
public 
    PersianNormalizationFilterFactory(
	IDictionary<string, string> args) 
	        : base(args){
	if (args.Count > 0){
		throw new System.ArgumentException(
			"Unknown parameters: " 
			+ args
		);
	}
}
\end{lstlisting}
\end{tabular}
\end{subfigure}
\rule{\linewidth}{0.6pt}
\caption{
Qualitative examples of source code translation between Java and C\# using PLBART.
}
\label{cs_java_example}
\end{figure*}